\documentclass{article}

\usepackage[preprint]{neurips_2021}

\usepackage[utf8]{inputenc} 
\usepackage[T1]{fontenc}    
\usepackage{hyperref}       
\usepackage{url}            
\usepackage{booktabs}       
\usepackage{amsfonts}       
\usepackage{nicefrac}       
\usepackage{microtype}      
\usepackage{xcolor}         
\usepackage{xspace}
\usepackage{amsmath}
\usepackage{amssymb}
\usepackage{amsthm}
\usepackage{graphicx}
\usepackage{subcaption}
\usepackage[title]{appendix}

\graphicspath{{Figures/}}

\newcommand{\RT}{$\mathbb{R}^{2}$\xspace}
\newcommand{\SORP}{$S^{1} \times \mathbb{R}^{+}$\xspace}

\newcommand{\RTSO}{$\mathbb{R}^{2} \times S^{1}$\xspace}
\newcommand{\RTRP}{$\mathbb{R}^{2} \times \mathbb{R}^{+}$\xspace}
\newcommand{\RTSORP}{$\mathbb{R}^{2} \times S^{1} \times \mathbb{R}^{+}$\xspace}

\title{Similarity Equivariant Linear Transformation of Joint Orientation-Scale Space Representations} 
\author{%
  Xinhua ~Zhang 
  \\
  Department of Computer Science\\
  University of New Mexico\\
  \texttt{xinhua@unm.edu} \\
  \And Lance R. Williams\\
  Department of Computer Science\\
  University of New Mexico\\
  \texttt{williams@cs.unm.edu} \\
}

\begin{document}

\maketitle

\begin{abstract}

Convolution is conventionally defined as a linear operation on functions of one or more variables which commutes with shifts.
Group convolution generalizes the concept to linear operations on functions of group elements representing more general geometric transformations and which commute with those transformations. Since similarity transformation is the most general geometric transformation on images that preserves shape, the group convolution that is equivariant to similarity transformation is the most general shape preserving linear operator. Because similarity transformations have four free parameters, group convolutions are defined on four-dimensional, {\it joint orientation-scale spaces.} Although prior work on equivariant linear operators has been limited to discrete groups, the similarity group is continuous. In this paper, we describe linear operators on discrete representations that are equivariant to continuous similarity transformation. This is achieved by using a basis of functions that is {\it joint shiftable-twistable-scalable.} These {\it pinwheel} functions use Fourier series in the orientation dimension and Laplace transform in the log-scale dimension to form a basis of spatially localized functions that can be continuously interpolated in position, orientation and scale. Although this result is potentially significant with respect to visual computation generally, we present an initial demonstration of its utility by using it to compute a shape equivariant distribution of closed contours traced by particles undergoing Brownian motion in velocity. The contours are constrained by sets of points and line endings representing well known bistable illusory contour inducing patterns. Prior computational models of the bistability of these patterns demonstrated that the alternative percepts correspond to local optima in particle speed (scale). We replicate these results using a recurrent neural network implementing similarity equivariant convolution in a finite pinwheel basis.
\end{abstract}

\section{Introduction}
\label{sec:intro}
Broadly speaking, human vision is characterized by two different abilities pertaining to two dimensional images of the three-dimensional world.
The first is the ability to recognize instances of objects of one or more predefined types.
The second is the ability to reconstruct representations of the three-dimensional world itself.
Because the visual computations underlying recognition and reconstruction rely on two-dimensional images as input, 
two-dimensional shape is fundamental to both.
Although two-dimensional shape is not (in general) preserved in images of the three-dimensional world transformed by egomotion, it is nevertheless critical that visual computations commute with shape-preserving geometric transformations. If they did not, then the results of recognition and reconstruction computations would depend arbitrarily on the position and orientation of the observer in the world.

The most fundamental visual computation that commutes with a shape-preserving transformation is convolution. Convolution is a linear operator that is translation {\it equivariant.} More specifically, if an input image is translated by some vector, then the output image is translated by the same vector but is otherwise unchanged. Although convolution commutes with translation, it does not commute with more general shape-preserving transformations. In particular, it does not commute with similarity transformation, which extends translation by adding rotation and dilation. Because rotation and dilation are both components of optical flow, it is critical that linear operators serving as modules in visual computations be equivariant to these shape-preserving transformations as well. 

The dual requirements of linearity and equivariance to shape-preserving transformations partly explains the structure of the early stages of the ventral pathway of the human visual system. This is because, absent symmetries in the operators, {\it i.e.,} {\it isotropies,} generalized convolutions based on geometric transformations can only be defined on spaces possessing the same dimensionality as the transformations themselves. 
The initial stages in the ventral pathway are the retina, lateral geniculate nucleus (LGN), and primary visual cortex (V1).
Neurons in the retina and LGN have rotationally symmetric receptive fields in a range of sizes that have been described as difference of Gaussians \cite{Rodieck1965} or Laplacians of Gaussians \cite{Marr1982}. It is conventionally assumed that these neurons perform a multi-scale analysis of the input image by convolving with a set of dilated impulse response functions. However, it is worth observing that the domain of the {\it scale-space} representation, ${\mathbb R}^2 \times \mathbb{R}^+$, produced by this process possesses the minimum number of dimensions (three) required by a similarity equivariant linear operator with an {\it isotropic} impulse response function.

Since Hubel and Wiesel\cite{HUBEL1962}, it has been known that the receptive fields of {\it simple cells} in primary visual cortex of humans, cats and monkeys systematically vary in position, orientation and scale.
Simple cell receptive fields are typically modeled as {\it Gabor functions}\cite{Daugman1985},
which are the products of oriented sine and cosine gratings
and two-dimensional Gaussians. The sine and cosine gratings provide selectivity in orientation while the Gaussian provides selectivity in scale.
Although it very likely performs other information processing functions,
the most basic is that primary visual cortex
computes an over-complete frame-based representation of the 
input image by means of wavelet transform. In engineering terms, the simple cells of visual cortex form a {\it joint orientation-scale space}.
We observe that the domain of the joint orientation-scale space representation, ${\mathbb R}^2 \times S^1 \times {\mathbb R}^+$ possesses the minimum number of dimensions (four) necessary to define {\it non-isotropic} linear operators that are equivariant to similarity transformation, the most general shape-preserving transformation of functions of position, orientation and scale.

Due to the continuous nature of similarity transformation, which translates, rotates and dilates by real valued vectors, angles and scale factors, similarity equivariant linear operators might be dismissed as mathematical idealizations. This is because visual computations implemented by physical systems can only transform finite representations.
However, under relatively weak assumptions, it is possible to define visual computations based on finite representations that, for all practical purposes, possess equivariance to continuous geometric transformations.

If a visual computation does not require rotation or dilation equivariance, but only translation equivariance, then discrete convolution is often good enough. This is the case in most applications of convolutional neural networks, for example. However, the addition of rotation and dilation significantly complicates the problem. The rotation of a discrete image by any angle not evenly divisible by $\pi/2$ requires continuous interpolation of the discretely sampled spatial variables. Dilation of a discrete image also requires interpolation of the spatial variables. In lieu of interpolation, one can assume that the image is periodic in the spatial domain and band-limited in the frequency domain. This property is called \textit{shiftability} \cite{Simoncelli1992}. A continuous function with this property can be represented as a finite Fourier series. Significantly, discrete convolution of the Fourier series representations implements continuous convolution.
In group theoretic terms, the group convolution of two functions of the continuous translation group is reduced to group convolution of two functions of a discrete subgroup.
 
Shiftability in space can be generalized to steerability in orientation because rotation is translation in a polar representation\cite{Freeman1991}. Steerable filters have been used to achieve continuous rotation equivariance in visual computations by \cite{Weiler2017,Worrall2017a,Cheng2018}. In this paper, the fundamental problem addressed is how to achieve equivariance to continuous shape preserving geometric transformations in visual computations operating on finite representations. Translation equivariant linear operators are generalized to similarity equivariant linear operators by generalizing convolution of functions of two spatial variables to group convolution of functions of the continuous similarity group. As with translation, under relatively weak additional assumptions, continuous convolutions can be reduced to discrete convolutions defined on subgroups.

Similarity transformation consists of translation, rotation and dilation.
A finite basis that spans a space of functions related by translation by real valued vectors is termed {\it shiftable.} The analogous properties for rotation and dilation (which are equivalent to shiftability in a log-polar representation) are termed \textit{steerability} and \textit{scalability}.
Functions of orientation are inherently periodic.
This is not true for functions of scale.
Ideally, a similarity equivariant visual computation would operate on a basis of functions parameterized by a finite discrete subgroup of continuous similarity transform. Although finite and discrete, this basis would span the space of functions related by continuous similarity transformation. Such a basis would be termed {\it jointly shiftable-steerable-scalable}. The main impediment to achieving this goal is the lack of periodicity in the scale dimension.

Because rotation and dilation become translations in log-polar coordinates, defining the basis functions in this coordinate system is potentially the most straightforward way to achieve joint steerability and scalability. Sadly, the fact that the log-scale dimension is not naturally periodic complicates matters. Making matters worse, periodicity in the spatial, orientation and log-scale dimensions would by itself be insufficient to achieve joint shiftability-steerability-scalability. Another assumption is required, namely, that the basis functions must be localized in the spatial dimensions, ${\mathbb R}^2$. Stated differently, they must be zero outside of a compact region of support. Yet it is a fundamental property of Fourier transforms that if a function is compact in the frequency domain, it cannot simultaneously be compact in the spatial domain. This necessitates the relaxation of the localization requirement in ${\mathbb R}^2$. One way to do this would be to use a 2D Gaussian as an envelope to approximate a basis function with a compact region of support. Although 2D Gaussians have been commonly used for this purpose \cite{Zweck2004,Weiler2017,Worrall2017a,Sosnovik2019}, they are not ideal when continuous dilation is the goal since Gaussians are not eigenfunctions of shift; continuous dilation of a 2D Gaussian can only be achieved by convolving with another 2D Gaussian.

The \textit{analytical Fourier-Mellin transform} (AFMT) \cite{Ghorbel1994} is a combination of Fourier series in orientation $S^1$ and Laplace transform in log-scale, ${\mathbb R}^+$. The Laplace transform is a generalization of the Fourier transform to complex frequency. The use of AFMT is convenient for several reasons. Firstly, the AFMT basis functions can be continuously rotated by shifting them in the orientation dimension, $S^1$. Secondly, they can be continuously dilated by shifting them in the log-scale dimension, ${\mathbb R}^+$. This is accomplished using the shift property of Laplace transform. Thirdly, the real part of the Laplace transform frequency is a polynomial envelope, \textit{e.g.} $\rho^{-1}$, where $\rho$ is a radial parameter. The use of this envelope approximately satisfies the localization constraint in ${\mathbb R}^2$ and simultaneously permits approximate shiftability without periodicity in ${\mathbb R}^+$. To implement this, the Laplace transform frequency is densely sampled along a vertical line in the complex plane. These samples represent harmonics of different frequencies within compact envelopes of constant shape.
Lastly, the use of the AFMT for these purposes is greatly facilitated by the fact that there is an analytical expression for the Fourier transform of functions represented in the AFMT basis. To summarize, in this paper, we propose a steerable, approximately shiftable and scalable finite basis for functions of ${\mathbb R}^2 \times S^1 \times {\mathbb R}^+$ derived from the AFMT. Additionally, we describe an efficient implementation of group convolution based on the continuous similarity group for functions represented in this basis. Although the basis is only approximately shiftable and scalable, it seems to be the best compromise possible given the conflicting requirements.

Most prior work on equivariance in visual computation has been concerned with visual recognition. In contrast, the focus of our work is visual reconstruction, the ability to construct representations of the three-dimensional world from one or more two-dimensional images. Visual reconstruction has both topological and metric components. {\it Perceptual completion} is the process of identifying the traces of the boundaries of visible and partially occluded surfaces in image-centered coordinates. So called {\it shape-from} processes operating within two-dimensional domains defined by these boundaries use shading, texture, motion and stereo to embed the surfaces in three-dimensional space by inferring depth and surface orientation.
 Often dismissed as optical illusions, illusory contours are (in fact) the inferred boundaries of surfaces with reflectance matching that of background surfaces. 
 Like all visual computations, the computational process responsible for the completion of the boundaries of visible and occluded surfaces must be equivariant to similarity transformation.
 
 It has been proposed that the probability distribution of boundary shape can be modeled by the trajectories of particles moving in directions undergoing Brownian motion \cite{Mumford1994}.
 Williams and Jacobs \cite{Williams1997_1} introduced the idea of a {\it stochastic completion field}, a distribution of trajectories satisfying pairs of position
 and direction constraints and showed how it could be computed using a group convolution with a Green's function for the random process.
 Williams and Thornber \cite{Thornber1997} generalized the stochastic completion field by showing that the distribution of trajectories following closed paths through subsets of position constraints was given by the solution of an eigenvector-eigenvalue problem.
 Additionally, they demonstrated scale equivariant computation of this distribution by maximizing the eigenvalue over particles of all speeds,
 and showed how the circle and square bistable percepts in Koffka cross illusory contour stimuli in human vision correspond to distinct local optima in the eigenvalue function of speed.
 This bistability is an epiphenomenon of the scale equivariant nature of the visual computation underlying perceptual completion.
 However, their computation was not implemented in a finite basis of 
 ${\mathbb R}^2 \times S^1 \times {\mathbb R}^+$ and did not use biologically plausible algorithms for the eigenvector-eigenvalue computation or optimization over particle speed.
 Zweck and Williams \cite{Zweck2004} showed how the eigenvector-eigenvalue problem could be computed in a Euclidean equivariant manner in a finite basis spanning ${\mathbb R}^2 \times S^1$.
 This was accomplished by numerical integration of a Fokker-Planck equation modeling the distribution of boundary shape.
 Yet they did not address the issue of scale equivariance since their {\it shiftable-twistable} basis functions only spanned ${\mathbb R}^2 \times S^1$.

The contributions of this paper can be summarized as follows. Firstly, a shiftable-twistable and approximately scalable basis of functions for ${\mathbb R}^2 \times S^1 \times {\mathbb R}^+$ is described. Secondly, an efficient implementation of group convolution based on the continuous similarity group is defined for functions represented in this basis. Finally, as an example application, we demonstrate continuous similarity equivariant computation of stochastic completion fields in finite representations. This application represents the logical culmination of a series of papers on this topic by simultaneously achieving the most general geometric equivariance possible while employing only brain-like representations and algorithms.

\section{Methods}
The space that corresponds to similarity transformation is a space of position, orientation and scale. 
Let $\mathbb{R}^2$ be the real plane, 
$S^{1}$ be the circle and $\mathbb{R}^{+}$ be the positive half of the real line.
It follows that
the space for 2D similarity transformation is \RTSORP. 
Let $\mathbb{C}$ be the complex plane, $\vec{x}, \vec{y} \in \mathbb{R}^{2}$, $\theta,\phi \in S^{1}$ 
and $r,\rho \in \mathbb{R}^{+}$, the group convolution 
between functions $f, g : \mathbb{R}^{2} \times S^{1} \times \mathbb{R}^{+} \to \mathbb{C}$ is
\begin{equation}
  \label{eq:group-convolution}
  \{f \circledast g\}(\vec{x},\theta,r) = \int_{{\mathbb R}^2 \times S^1 \times {\mathbb R}^+} f(\vec{y},\phi,\rho) \: \mathcal{T}_{(\vec{y}, \phi, \rho)} \{g\}(\vec{x},\theta,r) \: d \vec{y} \: d\phi \: d\rho
\end{equation}
where $\mathcal{T}_{(\vec{y}, \phi, \rho)}$ is the operator that translates, rotates, and dilates functions of \RTSORP. 
We assume that rotation and dilation always happen before translation in $\mathcal{T}$,
because translation does not commute with rotation and dilation.
Since $f\circledast g \neq g \circledast f$, group convolution is not commutative \cite{Cohen2016}.
For this reason, we call the function $f$ an input, 
and the function $g$ a filter. 
\subsection{Joint Orientation-Scale Space}
\begin{figure}[t]
  \centering
  \begin{subfigure}[t]{0.45\textwidth}
    \centering
    \includegraphics[width=0.9\linewidth]{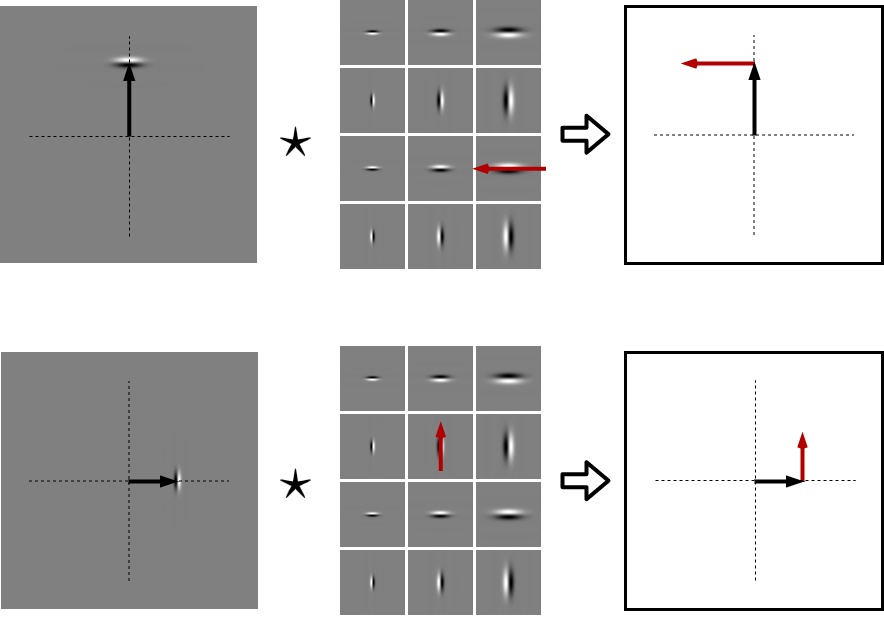}
    \caption{}
    \label{fig:GroupConvolutionIllustration}
  \end{subfigure}
  \begin{subfigure}[t]{0.45\textwidth}
    \centering
    \includegraphics[width=0.84\linewidth]{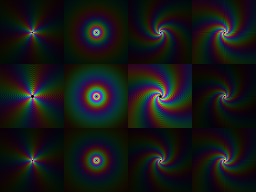}
    \caption{}
    \label{fig:pinwheel}
  \end{subfigure}  
  \caption{(a) Top row: An image that contains a canonical filter (left) cross-correlates with 
  the rotated and dilated canonical filters (middle).
  The position $(\phi, \rho, \theta, r)$ of the maximum response, 
  which corresponds to the filter in filter marked by a red vector, is displayed as two vectors (right). 
  The black vector corresponds to $(\phi, \rho)$ and the red vector corresponds to $(\theta, r)$. 
  Bottom row: The image is rotated and dilated, and the maximum response changes accordingly.
  Although we only show the transformation of the position of the maximum response, 
  the output of cross-correlation transforms in the same way. 
  (b) The phase of complex values are mapped to hue with red indicating positive real.
  For the first three columns from left,
  the bottom row consists of pinwheels defined by Eq. \eqref{eq:pinwheel_function}
  with frequencies $(\omega_{\phi}, \omega_{\rho}) = (5,0), (0,5)$ and $(5,5)$ , and $\alpha = -1$.
  They are plotted on a $64\times 64$ grid with grid width 1.
  The top row consists of Fourier pinwheels defined by Eq. \eqref{eq:AnalyticFourierPinwheel}
  with the same frequencies.
  The middle row consists of Fourier series computed using the Fourier pinwheels in the top row.
  The number of $\omega_{x}, \omega_{y}$ is $N = 256$,
  but the Fourier pinwheel is cropped to  $64\times 64$ for display purpose.
  The pinwheels in the right most column are Fourier series computed using Fourier pinwheels 
  with frequency $(\omega_{\phi}, \omega_{\rho}) = (5,5)$ and $N = 512,1024$ and $2048$ from top to bottom.}
  \label{fig:Local-Transformations}
\end{figure}

Although the space of functions is \RTSORP, a special case, 
\textit{i.e.}, $f: \mathbb{R}^{2} \to \mathbb{C}$, needs to be discussed.
In this case, the input $f$ consists of two separable functions: a function in $\mathbb{R}^{2}$ 
and a Dirac delta function in \SORP.
The filter $g$ consists of a set of rotated and dilated canonical filters in \RT, 
\textit{e.g.}, the middle column of Fig. \ref{fig:GroupConvolutionIllustration}.
It follows that the output of group convolution with such a filter, 
\textit{i.e.}, 
$(\circledast \: g) \in (\mathbb{R}^{2} \to \mathbb{C}) \to (\mathbb{R}^{2} \times S^{1} \times \mathbb{R}^{+} \to \mathbb{C})$, 
is four-dimensional.
This orientation and scale selective process produces representations in the joint orientation-scale space.
Accordingly, we call its subspaces, \RTSO and \RTRP, the orientation-space and scale-space,
respectively.
To distinguish the special case from the general one, we use the prime symbol to denote 2D function and transformation, \textit{e.g.} $f' \in \mathbb{R}^{2} \to \mathbb{C}$ and $f \in \mathbb{R}^{2} \times S^{1} \times \mathbb{R}^{+} \to \mathbb{C}$.

\subsection{Pinwheel Basis}
\label{sec:Pin-basis}
The use of log-polar coordinates is the key to combining rotation with dilation.
The Fourier transform in log-polar coordinates is the Fourier-Mellin transform (FMT).
Analytical Fourier-Mellin transform (AFMT) \cite{Ghorbel1994} generalized FMT by 
replacing the real frequency with a complex frequency in the scale dimension,
{\it i.e.,} Fourier transform is generalized to Laplace transform.
Since orientation is $2\pi$ periodic, the Fourier transform in orientation is actually a Fourier series. 
Let $\omega_{\phi} \in \mathbb{Z}$ be an angular frequency, 
$s = \alpha + i\omega_{\rho} \in \mathbb{C}$ be a radial frequency.
Let $(x,y)\in\mathbb{R}^{2}$ be Cartesian coordinates and
$\phi = \arctan{\frac{y}{x}}, \rho = \sqrt{x^2 + y^2}$ be the corresponding polar coordinates 
with origin $(0,0) \in \mathbb{R}^{2}$.
Finally, let $g' \in \mathbb{R}^{2} \to \mathbb{C}$ be a complex function
of two real variables.
The derivation of AFMT $\mathcal{M}'$ is in Appendix \ref{sec:analyt-four-mell-appendix}.
Because of the appearance of its basis functions
\begin{equation}
  \label{eq:pinwheel_function}
  h'(\phi,\rho,\omega_{\phi}, \alpha + i \omega_{\rho}) = e^{i\omega_{\phi}\phi}\rho^{\alpha + i \omega_{\rho}}
\end{equation}
\noindent we call this a \textit{pinwheel} basis (Fig. \ref{fig:pinwheel}).
Note that we use use polar coordinates here,
although AFMT is generally defined in log-polar coordinates.
Similarity transformation of the function $g'$ can be implemented using AFMT.
By using the shift properties of the Fourier and Laplace transforms, 
the function $g'$ can be continuously rotated and dilated, 
as the rotation and dilation in log-polar is a 2D translation.
In addition, the translation of the function $g'$ can be represented as translation in the pinwheel basis.
To continuously translate it, we have derived the Fourier transform of the pinwheel basis $h'$ in  Appendix \ref{sec:four-transf-pinwh-appendix}.
Since the pinwheel basis has an polynomial envelope, 
\textit{i.e.}. $\rho^{\alpha}$, it is effectively localized, and thus, 
the Fourier coefficients of a shiftable pinwheel can be derived from its Fourier transform 
by assuming periodicity of the spatial domain.
Let $P$ be the period of both $x$ and $y$ coordinates, 
$\omega_{x},\omega_{y}\in \mathbb{Z}$ be the corresponding Fourier coefficients.
Let $\bar{\phi} = \arctan\frac{\omega_{y}}{\omega_{x}}$ and
$\bar{\rho} = \sqrt{\omega_x^2 + \omega_y^2}$.
The Fourier coefficients of a shiftable pinwheel, comprising a \textit{Fourier pinwheel},
are given by
\begin{equation}
  \label{eq:AnalyticFourierPinwheel}
  \begin{split}
    H'(\omega_{\vec{x}},\omega_{\phi}, \alpha + i \omega_{\rho}) = &
    \begin{cases}
    \epsilon & \omega_{x} = \omega_{y} =0\\
    \pi (-i)^{|\omega_{\phi}|} \mathrm{e}^{j\omega_{\phi}\bar{\phi}}\left(\frac{P}{\pi \bar{\rho}}\right)^{^{2 + \alpha + i \omega_{\rho}}}  \frac{\Gamma\left (\frac{1}{2} (2 + |\omega_{\phi}| + \alpha + i\omega_{\rho})\right )}{\Gamma\left (\frac{1}{2}(|\omega_{\phi}| - \alpha - i\omega_{\rho})\right )} + \epsilon & \mathrm{otherwise}
  \end{cases}
  \end{split}
\end{equation}
\noindent where $\alpha \in (-2, -0.5)$, $\omega_{\vec{x}} = (\omega_{x}, \omega_{y})$ , 
$\Gamma$ is the Gamma function and $\epsilon$ is a constant that ensures the value at 
$(0,0) \in \mathbb{R}^{2}$ of the corresponding Fourier series is zero when 
it is represented discretely. 
Derivations can be found in Appendix \ref{sec:fourier-pinwheel-appendix}.
Despite its seeming complexity, the Fourier pinwheel is indeed a pinwheel in the frequency domain,
because the extra terms are constant with respect to $\omega_{\vec{x}}$ (Fig. \ref{fig:pinwheel}).

Let $\mathcal{F}'$ be the transform that computes 2D Fourier coefficients in \RT, which is the subspace of \RTSORP,
and $(\Delta \vec{x}, \Delta\theta, a) \in \mathbb{R}^{2} \times S^{1} \times \mathbb{R}^{+}$
be parameters for similarity transformation,
the 2D similarity transformation $\mathcal{T}'$ is defined by
\begin{equation}
  \label{eq:sim-trans-AFMT-R2}
  \begin{split}
    &\mathcal{T}'_{(\Delta \vec{x}, \Delta\theta, a)}\left\{g'\right\}\\
    &= \mathcal{F}'^{-1} \left\{\sum_{\omega_{\phi}}\int_{-\infty}^{\infty}\mathcal{M}'\{g'\}(\omega_{\phi},\alpha + i\omega_{\rho})H'(\omega_{\vec{x}},\omega_{\phi}, \alpha + i \omega_{\rho})a^{-(\alpha + i \omega_\rho)}e^{-i\left(\omega_{\phi}\Delta\theta  + \omega_{\vec{x}}\cdot\Delta\vec{x} \right)}\: d \omega_{\rho}\right\}
  \end{split}  
\end{equation}
\noindent where $(\cdot)$ is the inner-product.
Note that $\omega_{x}, \omega_{y}, \omega_{\phi} \in \mathbb{Z}$ are integers 
and $\omega_{\rho} \in \mathbb{R}$ is a real number.
Because the scale-dimension is non-periodic (at least the $\rho^{\alpha}$ envelope is non-periodic when $\alpha \neq 0$) , $\omega_{\rho}$ is not an integer.
To make the computation finite and discrete, $\omega_{\rho}$ is
densely sampled 
,
and the requirement that the pinwheel be effectively localized
necessitates the use of a $\rho^{\alpha}$ envelope with $\alpha < 0$.
It follows that the continuous similarity transformation defined in Eq. \eqref{eq:sim-trans-AFMT-R2} 
has been achieved at the expense of the completeness of the transform. 

\subsection{Similarity Equivariant Group Convolution}
\label{sec:Sim-Equi-G-Conv}
In this section, we use the transformation defined in Eq. \eqref{eq:sim-trans-AFMT-R2}
to implement continuous similarity equivariant group convolution.
For the special case we mentioned at the beginning of this section,
let $f' :\mathbb{R}^2 \to \mathbb{C}$ be a 2D function and
$g' : \mathbb{R}^{2} \to \mathbb{C}$ be a 2D canonical filter.
The input $f \in \mathbb{R}^{2} \times S^{1} \times \mathbb{R}^{+} \to \mathbb{C}$
consists of $f'$ and a Dirac delta function in \SORP.
The filter $g : \mathbb{R}^{2} \times S^{1} \times \mathbb{R}^{+} \to \mathbb{C}$
consists of rotated and dilated $g'$.
Let $\hat{\mathcal{M}}'$ represents the AFMT in \SORP, which is the subspace of \RTSORP,
the group convolution between $f$ and $g$ satisfies the following equation,
\begin{equation}
  \label{eq:GroupConvolutionR2}
    \mathcal{F}'\circ\hat{\mathcal{M}}'\left\{f \circledast g
    \right\}(\omega_{\vec{x}},-\omega_{\phi},-(\alpha + i\omega_{\rho}))
    = \mathcal{F}'\{f'\}(\omega_{\vec{x}})\mathcal{M}'\{g'\}(\omega_{\phi},\alpha + i\omega_{\rho})H'(\omega_{\vec{x}},\omega_{\phi}, \alpha + i \omega_{\rho})
\end{equation}
\noindent where $(\circ)$
is the function composition operator.
Derivations can be found in Appendix \ref{sec:group-conv-joint-sp-appendix}.
However, it can be understood as an application of the convolution theorems 
for Fourier and Laplace transforms.
We observe that the output of the group convolution is continuous in all four dimensions.

To derive the general case of the group convolution in the joint orientation-scale space, we need to define the transformation $\mathcal{T}$ in Eq. \eqref{eq:group-convolution} first. 
In particular, the difference between $\mathcal{T}$ and $\mathcal{T}'$ is the method used to rotate and dilate.
Translation is more straightforward, since given a representation in the basis,
a function can be translated simply by translating the pinwheel basis functions. 
Let $f, g \in \mathbb{R}^{2} \times S^{1} \times \mathbb{R}^{+} \to \mathbb{C}$ be the functions in 
the joint orientation-scale space, the rotation and dilation in this space is defined by
\begin{equation}
  \label{eq:Translation_Rotation_Scale_R2S1RP}
  \mathcal{T}_{(\vec{0}, \Delta\theta, a)}\left\{g\right\}(\phi, \rho, \theta, r) = g\left(\phi - \Delta\theta, \frac{\rho}{a}, \theta - \Delta\theta, \frac{r}{a}\right).
\end{equation}
The reason that the coordinates $(\phi, \rho)$ and $(\theta, r)$ 
are transformed in the same way is illustrated in Fig. \ref{fig:GroupConvolutionIllustration}.
Cross-correlation is used to illustrate the idea since it is more straightforward to visualize, yet convolution behaves the same way.
It follows that $(\phi, \rho, \theta, r)$ are not orthogonal coordinates with respect to the transformation $\mathcal{T}$.
The orthogonal coordinates that support the transformation most naturally are $(\phi, \rho , \theta - \phi, \frac{r}{\rho})$.
As shown in Fig. \ref{fig:GroupConvolutionIllustration}, $\theta - \phi$ is the angle between the red and black vectors, 
and $\frac{r}{\rho}$ is the ratio of their lengths.
Significantly, the transformation $\mathcal{T}$ does not change the shape of the parallelogram defined by the two vectors.
Since these four orthogonal coordinates span the space $(S^{1}\times\mathbb{R}^{+})^2$,
we can define AFMTs for the two \SORP subspaces.
Let $\omega_{\phi} \in \mathbb{Z}$, $\omega_{\theta} \in \mathbb{Z}$, 
$s_{\rho} = \alpha_{\rho} + i \omega_{\rho} \in \mathbb{C}$ and $s_{r} = \alpha_{r} + i \omega_{r} \in \mathbb{C}$
be integer and complex frequencies parameterizing the basis functions in the 
$\phi$, $\theta - \phi$, $\log{\rho}$ and $\log{\frac{r}{\rho}}$ axes respectively,
then the AFMT basis of the joint orientation-scale space $\mathcal{M}$ is defined by
\begin{equation}
  \label{eq:PinwheelR2S1R+}
  \begin{split}
    &\indent h(\phi, \rho, \theta, r, \omega_{\phi}, \alpha_{\rho}+i\omega_{\rho}, \omega_{\theta}, \alpha_{r} + i\omega_{r})
    =\rho^{\alpha_{\rho} + i\omega_{\rho}}e^{i\omega_{\phi}\phi}\left(\frac{r}{\rho}\right)^{\alpha_{r} + i\omega_{r}}e^{i\omega_{\theta}(\theta - \phi)}\\   
    &= h'(\phi, \rho,\omega_{\phi} - \omega_{\theta},\alpha_{\rho} - \alpha_{r} + i(\omega_{\rho} - \omega_{r}))h'(\theta, r, \omega_{\theta}, \alpha_{r} + i\omega_{r}).
  \end{split}
\end{equation}
We observe that the basis $h$ can be factored into two component pinwheel bases. It is a {\it joint pinwheel basis}. 
The domains of the position and velocity pinwheel bases functions are spanned by the 
polar coordinates $(\phi, \rho)$ and $(\theta, r)$
respectively.
Given a representation of a function of the joint orientation-scale space,
translation is accomplished by translating the coordinates
of the position basis functions alone, and it can be implemented by using Fourier pinwheel $H'$.
The transformation $\mathcal{T}$ represented using the AFMT in the joint orientation-scale space is
\begin{equation}
  \label{eq:SimilarityTransformationAFMTR2S1RP}
  \begin{split}
    &\mathcal{T}_{(\Delta \vec{x}, \Delta\theta, a)}\{g\}(x,y, \theta, r) = \\
    &\mathcal{F}'^{-1}\left\{
    \begin{split}
      \sum_{\omega_{\phi}, \omega_{\theta}}\iint_{-\infty}^{\infty}&\mathcal{M}\{g\}(\omega_{\phi},\alpha_{\rho} +  i\omega_{\rho},\omega_{\theta},\alpha_{r} + i\omega_{r})H'(\omega_{\vec{x}},\omega_{\phi} - \omega_{\theta},\alpha_{\rho} - \alpha_{r} + i(\omega_{\rho} - \omega_{r}))\\
    &h'(\theta, r, \omega_{\theta}, \alpha_{r} + i\omega_{r})e^{-i\left(\omega_{\phi}\Delta\theta  + \omega_{\vec{x}}\cdot\Delta\vec{x} \right)}a^{-(\alpha_{\rho} + i \omega_\rho)}\: d \omega_{\rho}\: d \omega_r
  \end{split}
  \right\}
  \end{split}. 
\end{equation}
The rotation and dilation in both $\mathbb{R}^2$ and \SORP are simultaneously accomplished. This is because changing $(\phi, \rho)$ of the orthogonal coordinates $(\phi, \rho , \theta - \phi, \frac{r}{\rho})$  does not change $(\theta - \phi, \frac{r}{\rho})$. Consequently, $(\theta, r)$ are changed by the same amount. The continuous group convolution in the joint orientation-scale space satisfies the following equation,
\begin{equation}
  \label{eq:GroupConvolutionR2S1R+}
  \begin{split}
    \mathcal{F}'\circ\hat{\mathcal{M}}' &\left \{f \circledast g\right \}
    (\omega_{\vec{x}}, \omega_{\theta}, \alpha_{r} + i\omega_{r}) = \\
    &\sum_{\omega_{\phi}, \omega_{\rho}}
    \mathcal{F}'\circ\hat{\mathcal{M}}' \left \{ f\right \}(\omega_{\vec{x}},\omega_{\phi},\alpha_{\rho} + i\omega_{\rho})
    \mathcal{M}\{g\}(\omega_{\phi},\alpha_{\rho} +  i\omega_{\rho},\omega_{\theta},\alpha_{r} + i\omega_{r})
     \\ &\qquad\ H'(\omega_{\vec{x}},\omega_{\phi} - \omega_{\theta}, \alpha_\rho - \alpha_r + i (\omega_{\rho} - \omega_{r})).
  \end{split}  
\end{equation}
Derivations can be found in Appendix \ref{sec:group-conv-joint-appendix}.
Note that the envelope parameter of the input function $f$ is $\alpha_{\rho}$, whereas the one of the output is $\alpha_{r}$. 
When they are not the same and the output of group convolution is used as an input of another group convolution, the mismatched envelope is a problem. 
It can be solved by spatially convolving the output of group convolution with a function that has a 2D Gaussian function in it, \textit{e.g.} the bias function in Appendix \ref{sec:bias-appendix}.


\section{Related Work on Group Equivariant Convolution}

Prior work on equivariance of group convolution can be categorized based on the set of geometric transformations implemented and whether the equivariance was continuous or discrete in its action in the spatial dimensions ${\mathbb R}^2$, orientation dimension $S^1$ and scale dimension $R^+$.
For example, Cohen and Welling \cite{Cohen2016} described group convolution for a subset of the similarity group including translation, rotation and reflection.
However, translation and rotation were only implemented for discrete subgroups
and rotation was limited to multiples of $\pi/2$ so that the problem of interpolation in ${\mathbb R}^2$ was completely avoided.
Worral et al. \cite{Worrall2017a} achieved continuous rotation equivariance in $\mathbb{R}^{2}$ by using circular harmonics although these were not used as basis functions. 
Marcos et al. \cite{Marcos2017} defined a vector field that was formed by the maximum response of the convolution of an input image with a discrete set of orientation tuned filters. 
The vector field can be viewed as a simplified orientation-space.
Weiler et al. \cite{Weiler2017} achieved continuous rotation equivariance in $\mathbb{R}^{2}$ using steerable filters as basis functions and created genuine orientation-space representations.
The group convolution is mixed discrete-continuous in the orientation-space since the action in ${\mathbb R}^2$ was continuous but the representation in $S^1$ was discrete and non-shiftable.
Weiler and Cesa \cite{Weiler2019} gave a general solution of the kernel space constraint for the Euclidean group $E(2)$ and its subgroups. 
Indeed, the work of \cite{Worrall2017a,Marcos2017,Weiler2017} can be considered to be special cases of this general framework.
However, dilation was not treated and translation in $S^1$ was restricted to the discrete subgroup $C_N$ in the examples used.
In contrast, our work implements continuous group action of $SO(2)$ on the orientation-space, $\mathbb{R}^{2} \times S^{1}$.
Cheng et al. \cite{Cheng2018} defined an actual continuous orientation-space representation and a joint steerable basis can be that continuously and simultaneously rotated in \RT and translated in $S^1$.
However, this basis was used to transform a discrete set of orientation tuned filters, making the orientation-space representation discrete in $S^1$.

Unlike the prior work on rotation equivariance cited above, 
where authors sometimes 
implemented continuous group action in ${\mathbb R}^2$ but discrete group 
action in ${\mathbb S}^1$,
the work on scale equivariance is uniformly discrete in its actions 
in both $\mathbb{R}^{2}$ and $\mathbb{R}^{+}$. 
For example,
Worrall and Welling \cite{Worrall2019a} achieved equivariance to a discrete subgroup of dilation by employing a discrete scale space.
Sosnovik et al. \cite{Sosnovik2019} achieved similar results by dilating steerable bases in $\mathbb{R}^{2}$.
Although the basis functions were steerable, they were not used to achieve rotation equivariance. 
Since all of the works mentioned above involved convolution in $\mathbb{R}^{2}$, they achieved either discrete translation and mixed discrete-continuous rotation equivariance, or discrete translation and discrete scale equivariance.
None of them achieved continuous translation equivariance using bases of shiftable functions.

\section{Experiments}
\subsection{Shiftable Pinwheel}
\label{sec:shiftable-pin-exper}
Fig. \ref{fig:pinwheel} shows examples of Fourier series computed from Fourier pinwheels defined in Eq. \eqref{eq:AnalyticFourierPinwheel}. By comparing them with pinwheels with the same angular and radial frequencies defined in Eq. \eqref{eq:pinwheel_function}, we can see that the phase of Fourier series are accurate, but the envelopes are not. The accurate angular phase indicates that the truncation of spatial Fourier coefficients does not have any negative effect on its steerability. The difference between the envelopes can be reduced by increasing the number of spatial frequencies, \textit{i.e.} $\omega_{x}$ and $\omega_{y}$. However, there will be too much computation if we use 2048 spatial frequencies in practice. Even though the envelopes of the Fourier series computed using 256 spatial frequencies are not very accurate, we found out that it was enough for producing closed illusory contours with correct shapes.

\subsection{Closed Illusory Contour}
\label{sec:clos-illus-cont}
\begin{figure}[t]
  \centering
  \includegraphics[width=\linewidth]{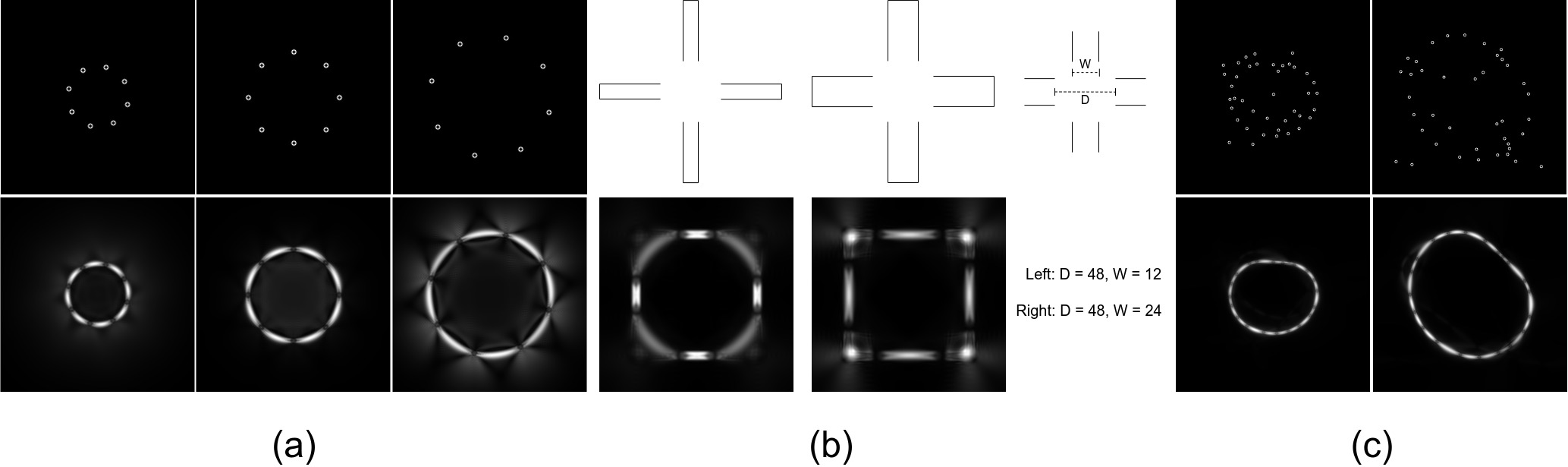}
  \caption{The representations in the joint orientation-scale space are plotted as 2D images by summing over the values with respect to \SORP. The points in the top row are inputs. The corresponding outputs are in the bottom row. (a) Koffka Cross. (b) Eight Dot Circle. (c) Avocado Shape with Noise.}
  \label{fig:Experiments}
\end{figure}

To demonstrate that we have achieved equivariance to continuous similarity transformation in a group convolution implemented in finite representations, we replicate the experiments described in Williams and Thornber \cite{Williams2001} and Williams and Zweck \cite{Williams2003}.
Williams and Thornber \cite{Williams2001} demonstrated scale equivariant computation of a distribution of particles undergoing Brownian motion in velocity following closed paths through subsets of given position direction pairs.
We refer to the first of their experiments as the ``eight dot circle'' since the input consisted of a set of 72 directions at each of 8 different positions on the circumference of a circle.
They demonstrated scale equivariant computation of a stochastic completion field dominated by particles tracing circular paths through the eight dots.
Neither particle speed, orientation or order of traversal were specified {\it a priori}.
The results they showed were computed using standard library implementations for eigenvector-eigenvalue decomposition and optimization of real valued functions.
We have replicated their result using repeated group convolution with a filter incorporating both the initial conditions and the Green's function for the same random process.
Combined with normalization, repeated convolution implements a power method iteration that converges to the same limiting distribution (eigenvector) as the computation described by Williams and Thornber.
Our work can therefore be understood as an implementation of the same computation but using brain-like representations and algorithms.
Continuous similarity group equivariance was essential to achieving this result since, without it, the recurrent neural network would not have converged to the correct limiting distribution. The details of experiment configuration are in Appendix \ref{sec:eight-dot-circle-appendix}. The results are plotted on a $256\times256$ grid with grid width 0.6 (Fig. \ref{fig:Experiments}a). The radii of circles from the left to the right are $24, 36$ and $48$. In addition, compared to the input of the circle in the middle, the ones of the circles on the left and right are rotated by 30 degree and then translated along $y$-coordinate by 0.7.

The Koffka cross is a well known illusory contour display exhibiting a bistable percept.
As the arms of the ``cross'' are widened, the shape of the perceived illusory contour changes from a more circular shape to a square shape.
Williams and Thornber \cite{Williams2001} argued that this transition was an epiphenomenon of scale equivariance since the circle and square percepts correspond to local optima in the eigenvalue function of particle speed.
As with the eight dot circle, the input consisted of a set of 72 directions at each of 8 different positions.
However, unlike the eight dot circle, the input was not isotropic,
since the intention was to represent the ends of the lines comprising the Koffka cross display and it is well known that illusory contours induced by line endings are predisposed to have orientations orthogonal to the lines.
As with the eight dot circle, we achieved the comparable results to Williams and Thornber but using brain-like algorithms and representations based on repeated similarity equivariant group convolution implemented in a finite pinwheel basis.
Significantly, the stochastic completion fields computed by our method display the same circle to square transition, but this occurs as a natural consequence of our use of scalable basis functions to represent the distribution of particle speeds, not by explicit optimization of the eigenvalue function of speed as was done in Williams and Thornber. The details of experiment configuration are in Appendix \ref{sec:koffka-cross-appendix}. The results are plotted on a $256\times256$ grid with grid width 0.3 (Fig. \ref{fig:Experiments}b).

Williams and Zweck \cite{Zweck2004} showed how the eigenvector-eigenvalue problem described by Williams and Thornber \cite{Williams2001} could be computed in a Euclidean equivariant manner in a finite basis of shiftable-twistable functions.
Our work differs from the work of Williams and Zweck in two ways.
The first is the manner in which the Fokker-Planck equation is solved.
The basic step in the Williams and Zweck power method involves integrating a Fokker-Planck equation using a finite difference method.
In contrast, our approach uses group convolution based on the similarity group with a Green's function.
These two ways are of course closely related since a
Green's function is the solution of a Fokker-Planck equation with a Dirac delta function initial condition.
General solutions of the Fokker-Planck equation can be computed by group convolution of of initial conditions consisting of superpositions of Dirac delta functions by group convolution with the Green's function.
The second way that our work differs from Williams and Zweck is that we achieve full similarity group equivariance because we address the problem of continuous scalability in finite representations.
This is a problem that they did not attempt to solve.

Williams and Zweck showed how a sparse set of dots on the boundary of a smooth closed contour could be reliably isolated from the dots of a random background pattern.
The Euclidean equivariance of this {\it perceptual saliency} computation was demonstrated by varying the position and orientation of the closed contour.
In our last experiment, we use the same scenario but demonstrate the similarity equivariance of our computation by varying the size of the closed contour,
in addition to its position and orientation.
The details of experiment configuration are in Appendix \ref{sec:avocado-shape-with-noise-appendix}. The results are plotted on a $256\times256$ grid with grid width 1 (Fig. \ref{fig:Experiments}c). 20 noise points are added to the 20 points that are sampled from the contour of an avocado. The input on the right uses the same noise sampling process, and then it is reflected, rotated, dilated and translated. The corresponding parameters are  10 degree (reflection angle),  70 degree (rotation angle), 1.5 (scale factor) and 0.7 (translation along $y$-axis).

\section{Conclusion}
\label{sec:conclusions}
In this paper, we introduced the idea of a joint orientation-scale space, ${\mathbb R}^2\times S^1 \times {\mathbb R}^+$, and defined the continuous similarity group action for this space. We describe a joint pinwheel basis for functions of this space consisting of pairs of polar-separable functions based on analytical Fourier-Mellin transform (AFMT).
The first pinwheels form a shiftable basis for ${\mathbb R}^2$ consisting of localized steerable-scalable functions.
The second pinwheels form a steerable-scalable basis for functions of $S^1 \times {\mathbb R}^+$.
Together, the joint pinwheel basis functions span the 
joint orientation scale space and
implement the continuous similarity group action on 
this space despite the finiteness of the basis.
This is (of course) accomplished at the expense of the completeness of the representation since the complex frequencies in the AFMT implementing scalability in both pinwheels are discretely sampled.
We described continuous similarity group equivariant convolution of functions of the joint orientation-space and showed how it could be implemented on finite representations in the joint pinwheel basis.
Finally, as an initial demonstration of the utility of our methods, we used them to compute a shape equivariant distribution of closed contours traced by particles undergoing Brownian motion in velocity.
This allowed us to replicate prior work on the phenomenon of illusory contours in human vision using brain-like representations and algorithms while demonstrating
equivariance to continuous similarity transformation.

\bibliographystyle{plain}
\bibliography{main}

\newpage
\appendix

\begin{appendices}
\section{Analytical Fourier-Mellin Transform}
\label{sec:analyt-four-mell-appendix}
Let $(x,y) \in \mathbb{R}^{2}$ be the Cartesian coordinates, $\phi = \arctan{\frac{y}{x}}$ and $\rho = \sqrt{x^{2} + y^{2}}$ be the corresponding polar coordinates, $\omega_{\phi} \in \mathbb{Z}$ be the angular frequency,  $s = \alpha + i\omega_{\rho} \in \mathbb{C}$ be the radial frequency  and $g' \in \mathbb{R}^{2} \to \mathbb{C}$, the combination of Fourier transform in $\phi$-axis and bilateral Laplace transform in $\log{\rho}$-axis is the analytical Fourier-Mellin transform $\mathcal{M}'$ \cite{Ghorbel1994}.  
\begin{equation}
  \label{eq:FourierTransformLogPolar}
  \begin{split}
    \hat{g}'(\omega_{\phi},\alpha + i \omega_{\rho})
    &= \mathcal{M}'\{g'\}(\omega_{\phi},\alpha + i\omega_{\rho}) \\
    &= \frac{1}{2\pi}\int_{-\infty}^{\infty}\int_{0}^{2\pi}g'(\phi,\log{\rho})e^{-i\omega_{\phi}\phi} e^{-(\alpha + i\omega_{\rho})\log{\rho}}\: d\phi\: d(\log{\rho})\\
    &= \frac{1}{2\pi}\int_{0}^{\infty}\int_{0}^{2\pi}g'(\phi,\rho)e^{-i\omega_{\phi}\phi}\rho^{-(\alpha + i\omega_{\rho})}\: d\phi\frac{\: d\rho}{\rho}
  \end{split}  
\end{equation}
The inverse transform involves a Mellin inversion, which is a line integral in the complex plane. As the integral is over a vertical line that passes through $\mathit{Re}\{s\} = \alpha$, it equals to a inverse Fourier transform multiplied with an envelope.
\begin{equation}
  \label{eq:InverseFourierTransformLogPolar}
  \begin{split}
    g'(\phi,\rho)
    &= \mathcal{M}'^{-1}\{\hat{g}'\}(\phi,\rho)\\
    &= \frac{1}{2\pi i}\int_{\alpha - i \infty }^{\alpha + i \infty}\sum_{\omega_{\phi} = -\infty}^{\infty}\hat{g}'(\omega_{\phi},\alpha, \omega_{\rho})e^{i\omega_{\phi}\phi}e^{s\log{\rho}}\: ds\\
    &= \int_{-\infty}^{\infty}\sum_{\omega_{\phi} = -\infty}^{\infty}\hat{g}'(\omega_{\phi},\alpha,\omega_{\rho})e^{i\omega_{\phi}\phi}e^{(\alpha + i\omega_{\rho})\log{\rho}}\: d\omega_{\rho}\\
    &= \int_{-\infty}^{\infty}\sum_{\omega_{\phi} = -\infty}^{\infty}\hat{g}'(\omega_{\phi},\alpha ,\omega_{\rho})e^{i\omega_{\phi}\phi}\rho^{\alpha + i\omega_{\rho}}\: d\omega_{\rho}
  \end{split}  
\end{equation}

\section{Fourier Transform of Pinwheel}
\label{sec:four-transf-pinwh-appendix}
\subsection{Fourier Transform  in Polar Representation}
Let $(x,y) \in \mathbb{R}^{2}$ be the Cartesian coordinates, $u,v \in \mathbb{R}$ be the spatial frequencies regarding $x$- and $y$-dimension and $f' \in \mathbb{R}^{2} \to \mathbb{C}$, the two dimensional Fourier transform $\mathcal{F}'$is
\begin{equation}
  \mathcal{F}\{f\}(u,v) = \int_{-\infty}^{\infty} \int_{-\infty}^{\infty} f(x,y) e^{-i 2\pi (ux + vy)} \: dx\: dy.
\end{equation}
The polar coordinates in the spatial domain and the frequency domain are  $\rho = \sqrt{x^2 +y^2}, \phi = \arctan \frac{y}{x}$ 
and $\bar{\rho} = \sqrt{u^2 + v^2}, \bar{\phi} = \arctan \frac{v}{u}$ respectively.
Since
\begin{equation}
\begin{split}
ux + vy &= \rho \bar{\rho} (\cos\phi\cos\bar{\phi} + \sin\phi\sin\bar{\phi})\\
&= \rho \bar{\rho} \cos(\phi - \bar{\phi}),
\end{split}
\end{equation}
and the small area represented by the differentials of variables $\rho$ and $\phi$ is $\rho\: d\rho\: d\phi$, the two dimensional Fourier transform in polar coordinates is
\begin{equation}
\label{eq:FourierTransformPolar} 
\mathcal{F}'\{f\}(\bar{\phi}, \bar{\rho}) =  \int_{0}^{2\pi}\int_{0}^{\infty} f(\rho, \phi) e^{-i 2\pi \rho \bar{\rho} \cos(\phi - \bar{\phi})}\rho\: d\rho\: d\phi.
\end{equation}

\subsection{Fourier Transform of Pinwheel}
\label{sec:four-transf-pinwh}
Let $\mathbb{Z}$ be the space of integer, $\omega_{\phi} \in \mathbb{Z}$ be the angular frequency, and $s = \alpha + i\omega_{\rho} \in \mathbb{C}$ be the radial frequency, a pinwheel basis is defined by
\begin{equation}
  \label{eq:pinwheel_function}
  h'(\phi,\rho,\omega_{\phi}, \alpha + i \omega_{\rho}) = e^{i\omega_{\phi}\phi}\rho^{\alpha + i \omega_{\rho}}
\end{equation}
The Fourier transform of the pinwheel basis using the polar representation in Eq. \eqref{eq:FourierTransformPolar} is
\begin{equation}
\begin{split}
  \mathcal{F}'\{h'\}(\bar{\phi},\bar{\rho}) &=  \int_{0}^{2\pi}\int_{0}^{\infty} \rho^{\alpha + i \omega_{\rho}} e^{i\omega_{\phi}\phi} e^{-i 2\pi \rho \bar{\rho} \cos(\phi - \bar{\phi})}\rho\: d\rho\: d\phi\\
  &=  \int_{0}^{\infty} \rho^{\alpha + i  \omega_{\rho}} \int_{0}^{2\pi}e^{-i
    \left( 2\pi \rho \bar{\rho} \cos(\phi - \bar{\phi}) - \omega_{\phi}\phi \right)} \: d\phi \rho \: d\rho
\end{split}
\label{eq:FourierTransformOfPinwheelPolar} 
\end{equation}

To compute Eq.\eqref{eq:FourierTransformOfPinwheelPolar}, we need Bessel's integral. Let $x\in \mathbb{R}$ be the variable and $n\in \mathbb{Z}$ is the order, the Bessel's integral is defined by
\begin{equation}
\label{eq:BesselIntegral} 
J_n(x) = \frac{1}{2\pi} \int_{-\pi}^{\pi} e^{i(x\sin\tau - n\tau)} \: d\tau.
\end{equation}
Note that in Eq. \eqref{eq:FourierTransformOfPinwheelPolar} there is a $\cos(\phi - \bar{\phi})$, whereas there is a $\sin\tau$ in the Bessel's integral. Thus, we can compute the integral by the change of variable.  Let $\tau = t - \frac{\pi}{2} $,
\begin{equation}
\label{eq:ModifiedBesselIntegral} 
\begin{split}
J_n(x) &= \frac{1}{2\pi} \int_{-\pi}^{\pi} e^{i(x\sin\tau - n\tau)} \: d\tau\\
&= \frac{1}{2\pi} \int^{\frac{3}{2}\pi}_{-\frac{1}{2}\pi} e^{i(x\sin(t - \frac{\pi}{2}) - n(t - \frac{\pi}{2}))} \: d(t - \frac{\pi}{2})\\
&= \frac{1}{2\pi} \int^{\frac{3}{2}\pi}_{-\frac{1}{2}\pi} e^{i(-x\cos t - n t)} e^{i\frac{\pi}{2}n} \: dt\\
&=\frac{i^n}{2\pi} \int^{2\pi}_{0} e^{-i(x\cos t + n t)}  \: dt.
\end{split}
\end{equation}
Let $t = \phi - \bar{\phi}$, by substituting the integral over $\phi$ in Eq. \eqref{eq:FourierTransformOfPinwheelPolar} with Eq. \eqref{eq:ModifiedBesselIntegral}, we have
\begin{equation}
\begin{split}
  \mathcal{F}'\{h'\}(\bar{\phi},\bar{\rho}) 
  &=  \int_{0}^{\infty} \rho^{\alpha + i  \omega_{\rho}} \int_{0}^{2\pi}e^{-i
    \left( 2\pi \rho \bar{\rho} \cos(\phi - \bar{\phi}) - \omega_{\phi}\phi \right)} \: d\phi \rho \: d\rho\\
  &=\int_{0}^{\infty} \rho^{\alpha + i  \omega_{\rho}} \int_{0}^{2\pi}e^{-i\left( 2\pi \rho \bar{\rho} \cos(\phi - \bar{\phi}) - \omega_{\phi}(\phi - \bar{\phi}) - \omega_{\phi}\bar{\phi}  \right) } \: d(\phi - \bar{\phi}) \rho \: d\rho\\
  &=\int_{0}^{\infty} \rho^{\alpha + i  \omega_{\rho}} \int_{0}^{2\pi}e^{-i \left( 2\pi \rho \bar{\rho} \cos{t} - \omega_{\phi}t - \omega_{\phi}\bar{\phi}  \right) } \: dt \rho  \: d\rho\\
  &=2\pi (\frac{1}{i})^{-\omega_{\phi}}e^{i\omega_{\phi}\bar{\phi}}\int_{0}^{\infty} \rho^{\alpha + i  \omega_{\rho}} J_{-\omega_{\phi}}(2\pi\rho \bar{\rho})\rho\: d\rho
\end{split}
\label{eq:FourierTransformOfPinwheelIntermediate1}
\end{equation}
By using the property of Bessel's function, 
\begin{equation} 
  J_{-n}(x) = (-1)^{n}J_{n}(x).
  \label{eq:Bessels-property}
\end{equation}
Eq. \eqref{eq:FourierTransformOfPinwheelIntermediate1} can be rewritten as
\begin{equation}
  \mathcal{F}'\{h'\}(\bar{\phi},\bar{\rho}) 
  =2\pi (-i)^{\omega_{\phi}}e^{i\omega_{\phi}\bar{\phi}}\int_{0}^{\infty} \rho^{\alpha + i  \omega_{\rho}} J_{\omega_{\phi}}(2\pi\rho \bar{\rho})\rho\: d\rho
\label{eq:FourierTransformOfPinwheelIntermediate2}
\end{equation}

To compute Eq. \eqref{eq:FourierTransformOfPinwheelIntermediate2} we need Hankel transform,
\begin{equation}
H_n(k) = \int_0^\infty f(\rho) J_n(k\rho) \rho \: d \rho.
\end{equation}
When $f(\bar{\rho}) = \bar{\rho}^s$, where $r\in\mathbb{R}^{+}$ and $s\in\mathbb{C}$, its Hankel transform \cite{Piessens2000} is 
\begin{equation}
\label{eq:HankelTransform} 
H_n(k) = \mathcal{H}\{f\}(k) = \frac{2^{s+1}}{k^{s+2}}\frac{\Gamma\left (\frac{1}{2}(2 + n + s)\right )}{\Gamma\left (\frac{1}{2}(n - s)\right )},
\end{equation}
where $\Gamma$ is the Gamma function and $Re(s) \in (-2,-0.5)$. By substituting the integral over $\rho$ in Eq. \eqref{eq:FourierTransformOfPinwheelIntermediate1} with Eq. \eqref{eq:HankelTransform}, 
and let $k = 2\pi \bar{\rho}$ and $s = \alpha + i\omega_{\rho}$, we have
\begin{equation}
\mathcal{F}'\{h'\}(\bar{\phi}, \bar{\rho}) = \frac{\pi (-i)^{\omega_{\phi}} \mathrm{e}^{j\omega_{\phi}\bar{\phi}}}{(\pi \bar{\rho})^{2 + \alpha + i \omega_{\rho}}}  \frac{\Gamma\left (\frac{1}{2} (2 + \omega_{\phi} + \alpha + i\omega_{\rho})\right )}{\Gamma\left (\frac{1}{2}(\omega_{\phi} - \alpha - i\omega_{\rho})\right )}.
\end{equation}
To avoid negative integers in the Gamma functions, we use the property in Eq. \eqref{eq:Bessels-property}, 
\begin{equation}
  \mathcal{F}'\{h'\}(\bar{\phi}, \bar{\rho}) = \frac{\pi (-i)^{|\omega_{\phi}|} \mathrm{e}^{j\omega_{\phi}\bar{\phi}}}{(\pi \bar{\rho})^{2 + \alpha + i \omega_{\rho}}}  \frac{\Gamma\left (\frac{1}{2} (2 + |\omega_{\phi}| + \alpha + i\omega_{\rho})\right )}{\Gamma\left (\frac{1}{2}(|\omega_{\phi}| - \alpha - i\omega_{\rho})\right )}.
  \label{eq:FourierTransformOfPinwheel}
\end{equation}

\section{Fourier Pinwheel}
\label{sec:fourier-pinwheel-appendix}
\subsection{Fourier Coefficients of Shiftable Pinwheel}
\label{sec:four-coeff-shift-pin}
Since the Fourier transform of pinwheel exists only when $\alpha \in (-2, -0.5)$ (see Appendix \ref{sec:four-transf-pinwh}), the envelope $\rho^{\alpha}$ makes pinwheel effectively localized. It follows that we can assume that a pinwheel is periodic in the spatial domain. Let $(x,y) \in \mathbb{R}^{2}$ be the Cartesian coordinates, $\mathbb{Z}$ be the space of integer, $\omega_{\phi} \in \mathbb{Z}$ be the angular frequency, $s = \alpha + i\omega_{\rho} \in \mathbb{C}$ be the radial frequency, $\omega_{x}, \omega_{y} \in \mathbb{Z}$ be the Fourier coefficients regarding $x$ and $y$ coordinates respectively and $P$ be the period of pinwheel in both $x$ and $y$ coordinates, the Fourier coefficients of a periodic pinwheel can be derived from Eq. \eqref{eq:FourierTransformOfPinwheelPolar} as
\begin{equation}
\begin{split}
  c(\omega_{x}, \omega_{y}, \omega_{\phi}, \alpha + i\omega_{\rho}) = \frac{1}{P} \int_{0}^{2\pi}\int_{0}^{\frac{P}{2}} \rho^{\alpha + i \omega_{\rho}} e^{i\omega_{\phi}\phi} e^{-i \frac{2\pi}{P} \rho \bar{\rho} \cos(\phi - \bar{\phi})}\rho\: d\rho\: d\phi,
\end{split}
\label{eq:FourierCoefficientsOfPinwheelPolar} 
\end{equation}
where $\rho = \sqrt{x^2 +y^2}, \phi = \arctan \frac{y}{x}$ are the polar coordinates in the spatial domain and $\bar{\rho} = \sqrt{\omega_x^2 + \omega_y^2}, \bar{\phi} = \arctan \frac{\omega_y}{\omega_x}$ are the polar coordinates in the frequency domain. Because of the effective locality, we can further assume that the integral over $\rho$ in Eq. \eqref{eq:FourierTransformOfPinwheelIntermediate1} can be approximated by 
\begin{equation}
  \int_{0}^{\frac{P}{2}} \rho^{\alpha + i  \omega_{\rho}} J_{\omega_{\phi}}(\frac{2\pi}{P}\rho \bar{\rho})\rho\: d\rho \approx \int_{0}^{\infty} \rho^{\alpha + i  \omega_{\rho}} J_{\omega_{\phi}}(\frac{2\pi}{P}\rho \bar{\rho})\rho\: d\rho.
\end{equation}
Thus, the Fourier Pinwheel, which is the Fourier coefficients of a shiftable pinwheel, can be derived using the same process as the one in Appendix \ref{sec:four-transf-pinwh}, 
\begin{equation}
  c(\omega_{x}, \omega_{y}, \omega_{\phi}, \alpha + i\omega_{\rho}) = \pi (-i)^{|\omega_{\phi}|} \mathrm{e}^{j\omega_{\phi}\bar{\phi}}\left(\frac{P}{\pi \bar{\rho}}\right)^{^{2 + \alpha + i \omega_{\rho}}}  \frac{\Gamma\left (\frac{1}{2} (2 + |\omega_{\phi}| + \alpha + i\omega_{\rho})\right )}{\Gamma\left (\frac{1}{2}(|\omega_{\phi}| - \alpha - i\omega_{\rho})\right )}.
  \label{eq:FourierPinwheel}
\end{equation}
\subsection{Shiftable Pinwheel With Zero Value at Origin}
\label{sec:shift-pinwh-with-appendix}
The Fourier pinwheel is a pinwheel in the frequency domain. As it exists only when $\alpha \in (-2, -0.5)$, its envelope in the frequency domain is $\bar{\rho}^{\beta} = \bar{\rho}^{-(2 + \alpha)}$, where $\beta \in (-1.5,0)$. The Fourier pinwheel is also effectively localized in the frequency domain, and thus, we can use $\omega_{x}, \omega_{y} \in \mathbb{Z}$ in a finite range, \textit{e.g.} $[-N\dots N]$, where $N$ is a non-negative integer. However, the truncation of the spatial frequencies makes the corresponding Fourier series an approximation to the real shiftable pinwheel. One side effect of the approximation is that the value at the center of its Fourier series, \textit{i.e.} the origin of log-polar, is non-zero. The value of a pinwheel at the origin is undefined, because the radial coordinate of the origin represented in lop-polar coordinate system is $-\infty$. Thus, we assume that the value at the origin is zero. If follows that we need to make sure that the values at the origins of both the Fourier pinwheel and its Fourier series are zeros. The definition of the Fourier pinwheel in Eq. \eqref{eq:FourierPinwheel} is redefined as
\begin{equation}
  c'(\omega_{x}, \omega_{y}, \omega_{\phi}, \alpha + i\omega_{\rho}) =
  \begin{cases}
    0 & \omega_{x} = \omega_{y} =0\\
    \pi (-i)^{|\omega_{\phi}|} \mathrm{e}^{j\omega_{\phi}\bar{\phi}}\left(\frac{P}{\pi \bar{\rho}}\right)^{^{2 + \alpha + i \omega_{\rho}}}  \frac{\Gamma\left (\frac{1}{2} (2 + |\omega_{\phi}| + \alpha + i\omega_{\rho})\right )}{\Gamma\left (\frac{1}{2}(|\omega_{\phi}| - \alpha - i\omega_{\rho})\right )} & \mathrm{otherwise}
  \end{cases}.  
  \label{eq:FourierPinwheelHole}
\end{equation}
Since the value of its Fourier series at the origin is the summation of all of its Fourier coefficients, we can make it zero by subtracting a constant $\epsilon$. Suppose $\omega_{x} \in [-N_{x} \dots N_{x}]$ and $\omega_{y} \in [-N_{y} \dots N_{y}]$, where $N_{x}, 
N_{y}$ are non-negative integers,
\begin{equation}
  \label{eq:Epsilon}
  \epsilon = \frac{-1}{(2N_{x} + 1)(2N_{y} + 1)}\sum_{-N_{x}}^{N_{x}}\sum_{-N_{y}}^{N_{y}} c'(\omega_{x}, \omega_{y}, \omega_{\phi}, \alpha + i\omega_{\rho}).
\end{equation}
The Fourier pinwheel is redefined as
\begin{equation}
  c''(\omega_{x}, \omega_{y}, \omega_{\phi}, \alpha + i\omega_{\rho}) =
  \begin{cases}
    \epsilon & \omega_{x} = \omega_{y} =0\\
    \pi (-i)^{|\omega_{\phi}|} \mathrm{e}^{j\omega_{\phi}\bar{\phi}}\left(\frac{P}{\pi \bar{\rho}}\right)^{^{2 + \alpha + i \omega_{\rho}}}  \frac{\Gamma\left (\frac{1}{2} (2 + |\omega_{\phi}| + \alpha + i\omega_{\rho})\right )}{\Gamma\left (\frac{1}{2}(|\omega_{\phi}| - \alpha - i\omega_{\rho})\right )} + \epsilon & \mathrm{otherwise}
  \end{cases}.  
  \label{eq:FourierPinwheelFinal}
\end{equation}

\section{The Group Convolution in the Joint Orientation-Scale Space}\subsection{A Special Case}
\label{sec:group-conv-joint-sp-appendix}
Let $(\vec{x}, \theta, r) \in \mathbb{R}^{2} \times S^{1} \times \mathbb{R}^{+}$ be the coordinates of the joint orientation-scale space, $f' :\mathbb{R}^2 \to \mathbb{C}$ be a 2D function and
$g' : \mathbb{R}^{2} \to \mathbb{C}$ be a 2D canonical filter.
The input $f \in \mathbb{R}^{2} \times S^{1} \times \mathbb{R}^{+} \to \mathbb{C}$
consists of $f'$ and a Dirac delta function $\delta(\phi, \rho - 1)$ in \SORP.
The filter $g : \mathbb{R}^{2} \times S^{1} \times \mathbb{R}^{+} \to \mathbb{C}$
consists of rotated and dilated $g'$. The group convolution between $f$ and $g$ is 
\begin{equation}
  \label{eq:GroupConvolutionR2-appendix}
  \begin{split}
   \{f \circledast g\} (\vec{x}, \theta, r)
   &= \int_{{\mathbb R}^2 \times S^1 \times {\mathbb R}^+} f(\vec{y},\phi,\rho) \: \mathcal{T}_{(\vec{y}, \phi, \rho)} \{g\}(\vec{x},\theta,r) \: d \vec{y} \: d\phi \: d\rho\\
   &= \int_{{\mathbb R}^2 \times S^1 \times {\mathbb R}^+}  f'(\vec{y})\delta(\phi,\rho - 1)\mathcal{T}_{(\vec{y}, \phi, \rho)} \{g\}(\vec{x},\theta,r) \: d \vec{y} \: d\phi \: d\rho\\
   &= \int_{{\mathbb R}^2}  f'(\vec{y})\mathcal{T}_{(\vec{y}, 0, 1)} \{g\}(\vec{x},\theta,r) \: d \vec{y}\\
  \end{split}  
\end{equation}
Since $g$ consists of rotated and dilated $g'$
\begin{equation}
  \label{eq:GroupConvolutionR2-R2S1RP-S2-appendix}
  \mathcal{T}_{(\vec{y}, 0, 1)} \{g\}(\vec{x},\theta,r)
  = \mathcal{T}'_{(\vec{y}, \theta, r)}\left\{g'\right\}(\vec{x}),
\end{equation}
where $\mathcal{T}'$ is the 2D similarity transformation. Thus, the integral over $\vec{y}$ is a 2D convolution. In addition, $\mathcal{T}'$ can be represented using AFMT,
\begin{equation}
  \label{eq:GroupConvolutionR2-TG-appendix}
  \begin{split}
    &\mathcal{T}'_{(\vec{y}, \theta, r)}\left\{g'\right\}(\vec{x})\\
    &= \mathcal{F}'^{-1} \left\{\sum_{\omega_{\phi}}\int_{-\infty}^{\infty}\mathcal{M}'\{g'\}(\omega_{\phi},\alpha + i\omega_{\rho})H'(\omega_{\vec{x}},\omega_{\phi}, \alpha + i \omega_{\rho})r^{-(\alpha + i \omega_\rho)}e^{-i\left(\omega_{\phi}\theta  + \omega_{\vec{x}}\cdot\vec{y} \right)}\: d \omega_{\rho}\right\}
  \end{split}  
\end{equation}
where $\mathcal{F}'$ is the transform that computes 2D Fourier coefficients with respect to $\vec{x}$, $H'$ is the Fourier pinwheel, $(\cdot)$ is the inner-product, $\omega_{x},\omega_{y}\in \mathbb{Z}$ is the spatial Fourier coefficients, $\omega_{\vec{x}} = (\omega_{x},\omega_{y})$, $\omega_{\phi} \in \mathbb{Z}$ is the angular frequency and 
$s = \alpha + i\omega_{\rho} \in \mathbb{C}$ is the radial frequency. The summation over $\omega_{\phi}$ and integral over $\omega_{\rho}$ together with the Fourier pinwheel $H'$ computes the inverse AFMT of $\mathcal{M}'\{g'\}(\omega_{\phi},\alpha + i\omega_{\rho})r^{-(\alpha + i \omega_\rho)}e^{-i\omega_{\phi}\theta}$. Alternatively, we can use $r^{-(\alpha + i \omega_\rho)}e^{-i\omega_{\phi}\theta}$ as the basis of the inverse AFMT.
\begin{equation}
  \label{eq:GroupConvolutionR2-TG-1-appendix}
  \begin{split}
    \mathcal{T}'_{(\vec{y}, \theta, r)}\left\{g'\right\}(\vec{x})
    = \mathcal{F}'^{-1} \circ \hat{\mathcal{M}}'^{-1}\left\{\mathcal{M}'\{g'\}(\omega_{\phi},\alpha + i\omega_{\rho})H'(\omega_{\vec{x}},\omega_{\phi}, \alpha + i \omega_{\rho})e^{-i\omega_{\vec{x}}\cdot\vec{y}}\right\}
  \end{split}  
\end{equation}
where $(\circ)$ is the function composition operator, $\hat{\mathcal{M}}'^{-1}$ computes the inverse AFMT with respect to $(\theta, r)$ using frequencies $(-\omega_{\phi}, -(\alpha + i \omega_{\rho}))$.
Substitute Eq. \eqref{eq:GroupConvolutionR2-TG-1-appendix} and Eq. \eqref{eq:GroupConvolutionR2-R2S1RP-S2-appendix} into Eq. \eqref{eq:GroupConvolutionR2-appendix},  the integral over $\vec{y}$ compute the Fourier coefficients of the function $f$ using the basis $e^{-i\omega_{\vec{x}}\cdot\vec{y}}$
\begin{equation}
  \label{eq:GroupConvolutionR2-Spatila-appendix}
  \begin{split}
   \{f \circledast g\} (\vec{x}, \theta, r)
   = \mathcal{F}'^{-1}\circ\hat{\mathcal{M}}'^{-1}\left\{ \mathcal{F}'\left\{f'\right\}(\omega_{\vec{x}}) \mathcal{M}'\{g'\}(\omega_{\phi},\alpha + i\omega_{\rho})H'(\omega_{\vec{x}},\omega_{\phi}, \alpha + i \omega_{\rho})\right\}
  \end{split}  
\end{equation}
where $\hat{\mathcal{M}}'$ represents the AFMT in \SORP, which is the subspace of \RTSORP. Move the inverse $\mathcal{F}'$ and the inverse AFMT $\hat{\mathcal{M}}'$ to the left-hand side of the equation,
\begin{equation}
  \label{eq:GroupConvolutionR2-Final-appendix}
  \begin{split}
   \mathcal{F}'\circ \hat{\mathcal{M}}'\left\{f \circledast g\right\} (\omega_{\vec{x}},-\omega_{\phi},-(\alpha + i\omega_{\rho}))
   = \mathcal{F}'\{f'\}(\omega_{\vec{x}})\mathcal{M}'\{g'\}(\omega_{\phi},\alpha + i\omega_{\rho})H'(\omega_{\vec{x}},\omega_{\phi}, \alpha + i \omega_{\rho})
  \end{split}  
\end{equation}

\subsection{The General Case}
\label{sec:group-conv-joint-appendix}
Let $(\vec{x}, \theta, r) \in \mathbb{R}^{2} \times S^{1} \times \mathbb{R}^{+}$ be the coordinates of the joint orientation-scale space  and $f, g \in \mathbb{R}^{2} \times S^{1} \times \mathbb{R}^{+} \to \mathbb{C}$ be the functions in 
the joint orientation-scale space. The group convolution between $f$ and $g$ is 
\begin{equation}
  \label{eq:GroupConvolutionR2S1R+-spatial-appendix}
  \begin{split}
    \left \{f \circledast g\right \}(\vec{x}, \theta, r) 
    &= \int_{{\mathbb R}^2 \times S^1 \times {\mathbb R}^+} f(\vec{y},\phi,\rho) \: \mathcal{T}_{(\vec{y}, \phi, \rho)} \{g\}(\vec{x},\theta,r) \: d \vec{y} \: d\phi \: d\rho\\    
  \end{split}  
\end{equation}
The transformation $\mathcal{T}$ can be implemented using AFMT in the joint orientation-scale space,
\begin{equation}
  \label{eq:SimilarityTransformationAFMTR2S1RP-appendix}
  \begin{split}
    &\mathcal{T}_{(\vec{y}, \phi, \rho)}\{g\}(\vec{x}, \theta, r) = \\
    &\mathcal{F}'^{-1}\left\{
    \begin{split}
      \sum_{\omega_{\phi}, \omega_{\theta}}\iint_{-\infty}^{\infty}&\mathcal{M}\{g\}(\omega_{\phi},\alpha_{\rho} +  i\omega_{\rho},\omega_{\theta},\alpha_{r} + i\omega_{r})H'(\omega_{\vec{x}},\omega_{\phi} - \omega_{\theta},\alpha_{\rho} - \alpha_{r} + i(\omega_{\rho} - \omega_{r}))\\
    &h'(\theta, r, \omega_{\theta}, \alpha_{r} + i\omega_{r})e^{-i\left(\omega_{\phi}\phi  + \omega_{\vec{x}}\cdot\vec{y} \right)}\rho^{-(\alpha_{\rho} + i \omega_\rho)}\: d \omega_{\rho} \: d \omega_{r}
  \end{split}
  \right\}
  \end{split}  
\end{equation}
Substitute Eq. \eqref{eq:SimilarityTransformationAFMTR2S1RP-appendix} to Eq. \eqref{eq:GroupConvolutionR2S1R+-spatial-appendix}, the integral in \RTSORP can be divided into two parts. $\mathcal{F}'$ computes the Fourier coefficients of the function $f$ in \RT, using the basis $e^{-i\omega_{\vec{x}}\vec{y}}$, and $\hat{\mathcal{M}}'$ computes the 2D AFMT in \SORP, using the basis $e^{-i\omega_{\phi}\phi}\rho^{-(\alpha_{\rho} + i \omega_\rho)}$.
\begin{equation}
  \label{eq:GroupConvolutionR2S1R+-AFMT-appendix}
  \begin{split}
    &\left \{f \circledast g\right \}(\vec{x}, \theta, r) \\
    &= \mathcal{F}'^{-1}\left\{
    \begin{split}
      \sum_{\omega_{\phi},\omega_{\theta}}\iint_{-\infty}^{\infty}&\mathcal{M}\{g\}(\omega_{\phi},\alpha_{\rho} +  i\omega_{\rho},\omega_{\theta},\alpha_{r} + i\omega_{r})H'(\omega_{\vec{x}},\omega_{\phi} - \omega_{\theta},\alpha_{\rho} - \alpha_{r} + i(\omega_{\rho} - \omega_{r}))\\
      &h'(\theta, r, \omega_{\theta}, \alpha_{r} + i\omega_{r})
      \int_{{\mathbb R}^2 \times S^1 \times {\mathbb R}^+}f(\vec{y}, \phi,\rho)e^{-i\left(\omega_{\phi}\phi  + \omega_{\vec{x}}\cdot\vec{y} \right)}\rho^{-(\alpha_{\rho} + i \omega_\rho)} \: d \vec{y} \: d\phi \: d\rho\: d \omega_{\rho} d \omega_{r}
  \end{split}
  \right\}\\
  &= \mathcal{F}'^{-1}
    \left\{
    \begin{split}
      \sum_{\omega_{\phi},\omega_{\theta}}\iint_{-\infty}^{\infty}&\mathcal{F}'\circ\hat{\mathcal{M}}'\{f\}(\omega_{\vec{x}}, \omega_{\phi}, \alpha_{\rho} + i\omega_{\rho})\mathcal{M}\{g\}(\omega_{\phi},\alpha_{\rho} +  i\omega_{\rho},\omega_{\theta},\alpha_{r} + i\omega_{r})\\
      &H'(\omega_{\vec{x}},\omega_{\phi} - \omega_{\theta},\alpha_{\rho} - \alpha_{r} + i(\omega_{\rho} - \omega_{r}))h'(\theta, r, \omega_{\theta}, \alpha_{r} + i\omega_{r})\: d \omega_{\rho} d \omega_{r}
  \end{split}
\right\}\\
  &= \mathcal{F}'^{-1}\circ\hat{\mathcal{M}}'^{-1}
    \left\{
    \begin{split}
      \sum_{\omega_{\phi}}\int_{-\infty}^{\infty}&\mathcal{F}'\circ\hat{\mathcal{M}}'\{f\}(\omega_{\vec{x}}, \omega_{\phi}, \alpha_{\rho} + i\omega_{\rho})\mathcal{M}\{g\}(\omega_{\phi},\alpha_{\rho} +  i\omega_{\rho},\omega_{\theta},\alpha_{r} + i\omega_{r})\\
      &H'(\omega_{\vec{x}},\omega_{\phi} - \omega_{\theta},\alpha_{\rho} - \alpha_{r} + i(\omega_{\rho} - \omega_{r}))d \omega_{\rho}
  \end{split}
\right\}\\
  \end{split}  
\end{equation}
where $\hat{\mathcal{M}}'^{-1}$ computes the inverse 2D AFMT using the basis $h'(\theta, r, \omega_{\theta}, \alpha_{r} + i\omega_{r})$. Move $\mathcal{F}'^{-1}\circ\hat{\mathcal{M}}'^{-1}$ to the left-hand side of the equation
\begin{equation}
  \label{eq:GroupConvolutionR2S1R+-AFMT-Final-appendix}
  \begin{split}
    &\mathcal{F}'\circ\hat{\mathcal{M}}'\left \{f \circledast g\right \}(\omega_{\vec{x}}, \omega_{\theta}, \alpha_{r} + i \omega_{r}) \\
    &= 
    \left\{
    \begin{split}
      \sum_{\omega_{\phi}}\int_{-\infty}^{\infty}&\mathcal{F}'\circ\hat{\mathcal{M}}'\{f\}(\omega_{\vec{x}}, \omega_{\phi}, \alpha_{\rho} + i\omega_{\rho})\mathcal{M}\{g\}(\omega_{\phi},\alpha_{\rho} +  i\omega_{\rho},\omega_{\theta},\alpha_{r} + i\omega_{r})\\
      &H'(\omega_{\vec{x}},\omega_{\phi} - \omega_{\theta},\alpha_{\rho} - \alpha_{r} + i(\omega_{\rho} - \omega_{r}))d \omega_{\rho}
  \end{split}
\right\}\\
  \end{split}  
\end{equation}

\section{Bias}
\label{sec:bias-appendix}
A bias is the distribution regarding a position in $\mathbb{R}^{2}$ in the joint orientation-scale space. In order to be used in the proposed group convolution, it needs to be shiftable, steerable and scalable. In addition, it should be easily modified so as to add or remove orientation (or scale) preference. Finally, it needs to be able to solve the envelope mismatched problem, as the algorithm used to compute closed illusory contour problem is iterative. The output of group convolution will be multiplied with the bias in the spatial domain, and then be used as the input of group convolution.   A bias that meets these requirements is defined as follows. Let $\sigma_{\rho},\sigma_{r} \in \mathbb{R}$ be standard deviation of a Gaussian function in $\mathbb{R}^{2}$ and a Gaussian function in $\log{r}$ coordinate respectively, and $\gamma \in \mathbb{R}$ be parameter that controls the shape of polynomial envelope, the bias is defined by 
\begin{equation}
  \label{eq:bias}
  b(\phi, \rho, \theta, r) = e^{-\frac{\rho^{2}}{2\sigma_{\rho}^{2}}}\rho^{\gamma}e^{-\frac{(\log{r} - \log{\rho})^{2}}{2\sigma_{r}^{2}}}
\end{equation}
The physical meaning of the scale preference, \textit{i.e.} the Gaussian function in $\log{r}$, is that when a particle starts from one point, the further its initial position is away from the center of the point, the faster it travels. We design it to have a scale preference because it can avoid the self-loop problem \cite{Williams2003}. The Gaussian and polynomial envelopes in \RT provide effective locality and avoid the envelope mismatch problem. Additionally, we can derive an analytical expression of this bias by using the same trick we used on the Fourier pinwheel.
\section{Experiments Configurations}
\subsection{Eight Dot Circle}
\label{sec:eight-dot-circle-appendix}
The bias used to model the distribution of point for this problem is the one defined in Eq. \eqref{eq:bias}. The parameters for computing $G$ are $T = 0.018$ and $\tau = 9$. $G$ is sampled on a $512 \times 512 \times 90 \times 1$ grid. There are $512$ samples in each of $x$ and $y$ coordinates, 90 orientations (4 degree interval) and 1 scale. The spatial distance between adjacent sample points is $0.25$. The number of orientation and scale frequencies of pinwheels,  \textit{i.e.} $\omega_\phi, \omega_\rho, \omega_\theta$ and $\omega_r$, is 32. The angular frequency ranges in $[-16\dots15]$ and the scale frequency ranges in $[-8, -7.5 \dots 7.5]$. The envelope parameters of pinwheels are $\alpha_{\rho} = -1$ and $\alpha_{\rho} = 0$. The number of spatial frequencies, \textit{i.e.} $\omega_x$ and $\omega_y$ is 256, and the frequency ranges in $[-128 \dots 127]$. The period of both $x$ and $y$ dimension is $P = 256$. The parameters for the bias are $\sigma_\rho = 0.5, \sigma_r = 0.5$ and $\gamma = 10$. The number of iteration of the iterative algorithm is 10.

\subsection{Koffka Cross}
\label{sec:koffka-cross-appendix}
The bias that is used to model the endpoints of Koffka cross needs orientation preference. The orientation that is perpendicular to the one of the arm of Koffka cross is preferred. Let $\theta_a$ be the orientation of an arm of Koffka cross and $\sigma_\theta = 0.3$ be the standard deviation of the Guassuain function in coordinate, the bias is defined using the one in Eq. \eqref{eq:bias}
\begin{equation}
    b(\phi, \rho, \theta, r) \left(e^{-\frac{(\theta - (\theta_a + 0.5\pi))^{2}}{2\sigma_\theta}^2} + e^{-\frac{(\theta_a + (\theta_a + 0.5\pi))^{2}}{2\sigma_\theta}^2}\right)
\end{equation}
The Green's function $G'$ consists of the one regarding smooth contour, $G$, and the one regarding corner, $G_{c}$ \cite{Thornber1997}. The parameters for computing $G$ are $T = 0.016$ and $\tau = 9$.  $G' = G + \omega G_{c}$, where the weight $\omega = 0.000015$. Most of the rest parameters are the same to the ones in the eight dot circle experiment (see Appendix \ref{sec:eight-dot-circle-appendix}), expect that $\sigma_{r} = 0.75$.

\subsection{Avocado Shape with Noise}
\label{sec:avocado-shape-with-noise-appendix}
The parameters used in this experiment are the same to the ones in the eight dot circle experiment (see Appendix \ref{sec:eight-dot-circle-appendix}), expect that the number of iteration is 100.

\end{appendices}
\end{document}